 \documentclass[pmlr,twocolumn,10pt]{jmlr} 

\usepackage{booktabs}
\usepackage{siunitx}
\usepackage{arydshln}
\usepackage{svg}

\usepackage[switch]{lineno}

\newcommand{\equal}[1]{{\hypersetup{linkcolor=black}\thanks{#1}}}

\theorembodyfont{\upshape}
\theoremheaderfont{\scshape}
\theorempostheader{:}
\theoremsep{\newline}

\jmlrvolume{297}
\jmlryear{2025}
\jmlrworkshop{Machine Learning for Health (ML4H) 2025} 

 \title[A Bayesian Model for Multi-stage Censoring]{A Bayesian Model for Multi-stage Censoring}

\author{%
\Name{Shuvom Sadhuka} \Email{shuvom@csail.mit.edu}\\
\addr MIT
\AND
\Name{Sophia Lin}\Email{sophia\_lin@college.harvard.edu} \\
\addr Harvard University
\AND
\Name{Bonnie Berger}\equal{Co-last}\Email{bab@csail.mit.edu} \\
\addr MIT
\AND
\Name{Emma Pierson}\footnotemark[1] \Email{emmapierson@berkeley.edu}\\
\addr UC Berkeley
}

\begin{document}

\maketitle

\begin{abstract}
Many sequential decision settings in healthcare feature \textit{funnel} structures characterized by a series of stages, such as screenings or evaluations, where the number of patients who advance to each stage progressively decreases and decisions become increasingly costly. For example, an oncologist may first conduct a breast exam, followed by a mammogram for patients with concerning exams, followed by a biopsy for patients with concerning mammograms. A key challenge is that the ground truth outcome, such as the biopsy result, is only revealed at the end of this funnel. The selective censoring of the ground truth can introduce statistical biases in risk estimation, especially in underserved patient groups, whose outcomes are more frequently censored. We develop a Bayesian model for funnel decision structures, drawing from prior work on selective labels and censoring. We first show in synthetic settings that our model is able to recover the true parameters and predict outcomes for censored patients more accurately than baselines. We then apply our model to a dataset of emergency department visits, where in-hospital mortality is observed only for those who are admitted to either the hospital or ICU. We find that there are gender-based differences in hospital and ICU admissions. In particular, our model estimates that the mortality risk threshold to admit women to the ICU is higher for women (5.1\%) than for men (4.5\%).
\end{abstract}

\paragraph*{Data and Code Availability}
The data used in this study is sourced from the MIMIC-IV database, which requires credentialed access, which is granted after the completion of a data use agreement and required training on data security and patient privacy. Further details on accessing the MIMIC-IV database can be found at \href{https://mimic.mit.edu}{this link}. We release the code for our Stan model and preprocessing scripts at \href{https://github.com/shuvom-s/funnel_model/tree/main}{this link}. 

\paragraph*{Institutional Review Board (IRB)}
Our research does not require IRB approval, as MIMIC contains de-identified patient records. 

\section{Introduction}
\label{sec:intro}
Many important sequential decision-making settings follow a \textit{funnel} structure, where each successive stage involves fewer candidates or items yet brings additional information \citep{jain2023antiracist}. For instance, in healthcare, a clinician trying to diagnose a breast cancer may (1) conduct a breast exam, then (2) order a mammogram if the breast exam is concerning, then finally (3) order a biopsy if the mammogram is concerning \citep{johnshopkins_breast_ordering}. Crucially, fewer patients progress through each successive stage, and the ground truth outcome (biopsy result) is only observed at the final stage of this decision pipeline.

Funnel structures are a variant of the \textit{selective labels} problem. The selective labels problem describes settings in which ground truth outcomes (e.g., disease diagnosis) are \textit{censored} for a subset of the population as a result of a human decision (e.g., whether to test the patient for the disease). This selective censoring has been studied in healthcare \citep{mullainathan2022diagnosing}, lending \citep{fu2021crowds, guerdan2023ground}, and other high-stakes domains \citep{kleinberg2018human}. 

\begin{figure*}[ht] 
\centering
\setlength{\belowcaptionskip}{-2em} 
\includegraphics[
  width=0.9\textwidth,
  trim={1cm 4cm 1cm 2cm}, 
  clip
]{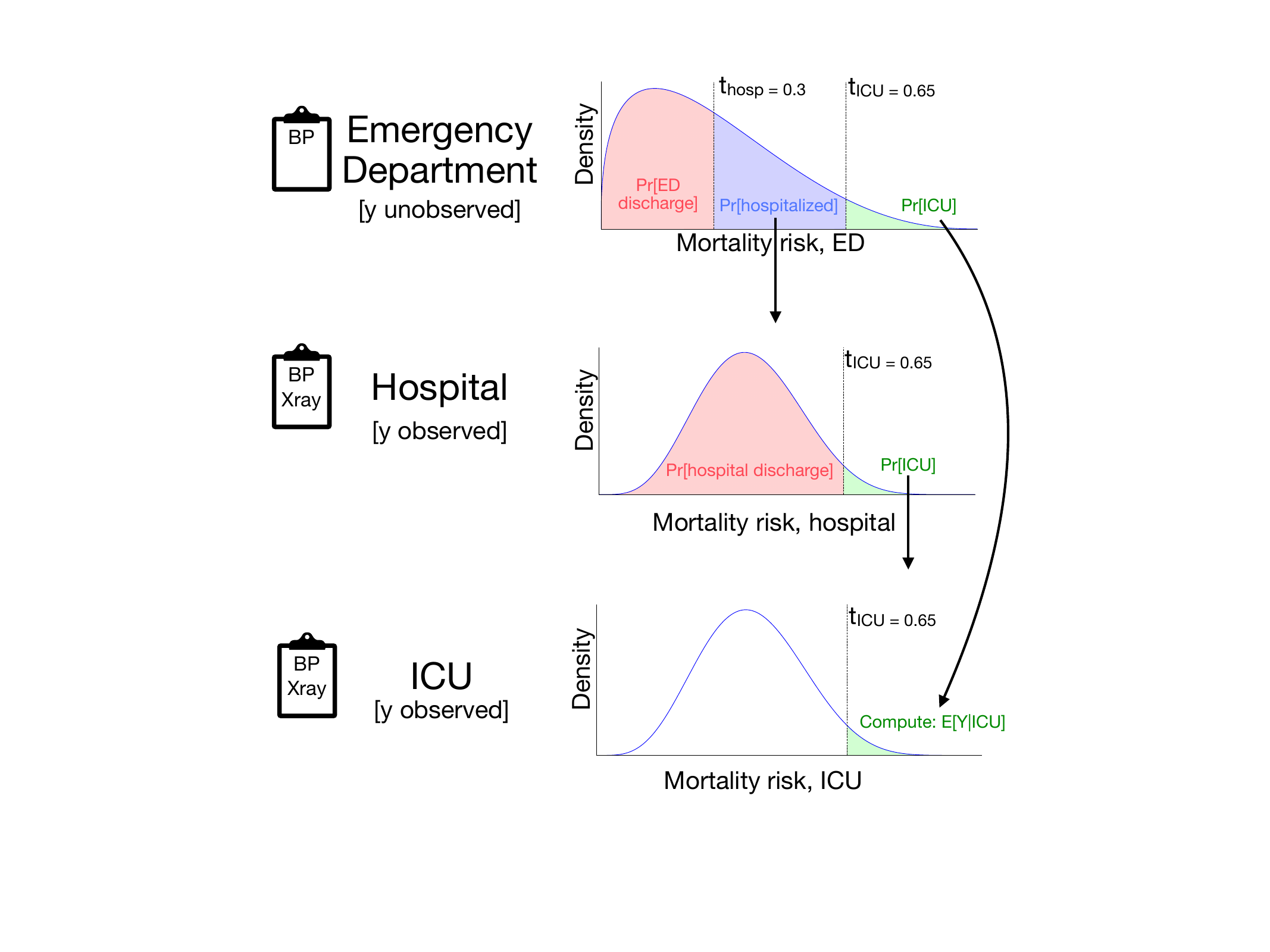}
\caption{Mechanics of the model. (Left) At each stage, the decision-maker estimates the mortality risk of a patient, given some covariates (e.g., blood pressure, or BP). If the drawn risk is larger than the threshold to hospitalize (or admit to ICU), the decision-maker moves them onto the appropriate stage. New variables, such as X-rays, may be collected at later stages, which are used to update the decision-maker's risk estimate. The ground truth is only observed for patients who are admitted to the hospital or ICU. (Right) A graphical representation of the patient flows. At each stage, $X$ represents the set of covariates available (e.g., BP and X-ray), while $D$ is the decision made for that patient at a particular stage (e.g., $D_{ED}$ could be discharge, admit to hospital, or admit to ICU). $Y$ is unobserved for patients discharged from the ED.}
\label{fig:model_mechanics}
\end{figure*}

There are risks in failing to account for this selective censoring. Any models fit only on patients whose ground truth outcomes are observed (e.g., disease diagnosis) may be statistically biased and fail to generalize to the population as a whole. Biases in early decisions — such as which patients are tested for a disease — can propagate throughout the care pipeline, producing systemic under-diagnoses in underserved populations, which may go undetected unless the selective censoring is modeled.

While the selective labels problem has been studied in settings with single decisions, we generalize this setup to sequences of nested (``funnel") decisions in the presence of unobserved covariates that affect both the human decision and the label (e.g., patient's visible discomfort). Additional examples of funnel structures outside healthcare include hiring, where a candidate must pass progressively harder interviews to be hired \citep{sackett1996multi}, lending, where an applicant must pass through several screenings to be approved for a loan \citep{chakravarty2009multistage}, and drug discovery, where a candidate molecule must pass through several phases of trials before FDA approval.

We introduce a Bayesian model for such decision \textit{funnels}. Our model jointly learns to predict both the ground truth outcomes and the human decisions (which selectively censor the ground truth). We generalize prior work on threshold tests \citep{pierson2018fast, simoiu2017problem}, originally developed in the context of policing, which explicitly account for unobserved covariates that influence both the human decisions (to censor a label) and the ground truth outcome. Specifically, threshold tests model the decision to test a patient (e.g., to administer a biopsy) as a function of some estimated risk (e.g., risk of actually having a cancer). If the estimated risk exceeds a decision threshold, the label is observed. We extend this test to funnel structures, where censoring occurs at multiple stages.

We first validate our model on synthetic data. We demonstrate, across a wide range of parameter settings, that our model is able to estimate the true parameters more accurately than binary classification baselines. In addition, our model estimates risk more accurately on the entire population (both censored and uncensored) than other binary classification baselines that do not account for multi-stage censoring.

We then apply this model to analyze real-world emergency department visits in MIMIC-IV, a Boston-area electronic medical record database. In particular, we focus on two critical emergency department decisions: (1) admitting a patient to the hospital, and (2) admitting a patient to the ICU. We consider in-hospital mortality to be the ground truth outcome of interest, which is only observed if the patient is admitted to the hospital or ICU. In contrast, if the patient is discharged from the ED, their ground truth is censored. At each stage of this funnel, patients are prioritized according to the urgency of care. We find that (1) men are admitted to the hospital and ICU at lower mortality risk thresholds than are women, and (2) men are assigned more severe triage acuities (assigned by a clinician) at the same estimated mortality risk than women, a finding that agrees with prior work on gender disparities in emergency and intensive care medicine \citep{todorov2021gender, patel2024patient}. We emphasize, however, that our model cannot explain the source of these gender-based disparities, and that the disparities, though statistically significant, are small in magnitude.

Overall, our contributions are as follows:

\begin{enumerate}   
    \item We generalize the \textit{selective labels} setting to a general setting with multi-stage censoring in the presence of unobserved covariates that influence both the human decisions and the label itself. We introduce a model to study multi-stage censoring in the presence of such unobservables by jointly modeling the true label and the human decisions.
    \item We show, in synthetic simulations, that our model outperforms baselines which do not account for this multi-stage censoring.
    \item We apply our model to real-world emergency department data from MIMIC-IV. We find that there are gender-based differences in hospital and ICU admissions decisions. Our model estimates that female patients must have a higher mortality risk (5.1\%) to be admitted to the ICU than male patients (4.5\%). 
\end{enumerate}
\section{Related Work}
\label{sec:relwork}
Several prior works learn decision policies for sequential decision tasks, including funnel-like structures \citep{jin2023selection}. At its core, funnel structures are a multi-stage extension of the selective labels problem \citep{lakkaraju2017selective}. In funnel structures, patients are given a label only if they pass several decision stages. \cite{jia2024learning} attempts to learn funnel decision policies which satisfy group fairness (i.e. some notion of parity is achieved at each stage). \cite{grossman2023racial} discusses the heterogeneity across decision makers in funnel structures and its relation to racial bias in pretrial detentions. \cite{wan2018item} develops a method for learning from sparse and noisy feedback from funnel structures in consumer recommender systems. Unlike our work, these works do not account for unobserved covariates that influence both the intermediate decisions and the label.

Other works have focused on developing models for sequential decision making more broadly. Within the context of healthcare, prior works have separately analyzed ED to hospitalization admissions \citep{barak2017early} and ICU admissions \citep{raita2019emergency}. Some works have focused on other, non-funnel sequential decision pathways in healthcare, including chronic disease management \citep{denton2018optimization} and surgery planning \citep{gul2015progressive}. 

Emergency department visits and decisions, in particular, are also valuable for other analyses. Prior literature has used ED data to monitor disease outbreaks \citep{wieland2007automated}, evaluate health risks \citep{patterson2019training}, assessed the expertise that doctors can provide over algorithms \citep{alur2024auditing}, and assessed racial disparities in algorithmic performance \citep{movva2023coarse}.

\section{Methods}
\label{sec:methods}

\subsection{Setup}
For concreteness, we describe our model in the setting of ED decision-making, but our approach can be applied to other funnel settings such as cancer screening or hiring. Let there be $N$ patients and $K$ decision stages. For example, if the funnel is ED $\rightarrow$ hospital $\rightarrow$ ICU, then $K=3$. For each patient $i$ who makes it to stage $k \in \{1, 2, \ldots, K\}$, the decision-maker observes a covariate set $X_{i,k}$, where covariates at earlier stages are (non-strict) subsets of covariates at deeper stages, i.e., $X_{i,1} \subseteq X_{i,2} \cdots X_{i,K}$. Let $s_i$ be the deepest stage patient $i$ progresses to. If $s_i \geq S$ for some pre-determined stage $S$ (e.g. hospital stage), the binary ground truth $y$ (e.g., in-hospital mortality) is revealed. Otherwise, $y$ is unobserved.

Patients need not progress through every stage sequentially. For instance, a patient with severe complications in the ED might be directly admitted to the ICU, bypassing hospitalization.

\textbf{Observed data.} Consider a particular patient $i$ progressing through different stages of a hospital stay (e.g., ED $\rightarrow$ hospital admission  $\rightarrow$ ICU). At each stage $k$, the clinician decides to move the patient on to a later stage or to discharge. We define the \textit{admission decision} at stage $k$ for patient $i$ as $D_{i,k}$, which takes on values in $\{ \text{discharge}, k+1, k+2, \ldots, K\}$. Note that $D_{i,k}$ can take on values larger than $k+1$, as the clinician can move the patient onto any later stage.

We additionally observe the patient's \textit{mortality} $Y_i \in \{0,1\}$ at discharge if the stage $k > S$. If the patient is discharged before stage $S$, $Y_i$ is unobserved in our data. Our observed data for each patient $i$ consists of a sequence of covariates $X_{i,k}$, admissions decisions $D_{i,k}$, and mortality $Y_i$ if the patient makes it past stage $S$. Otherwise, $Y_i$ is unobserved.

\subsection{Generative Model}
\textbf{Model.} We propose a generative model that extends prior work on \textit{threshold tests} \citep{simoiu2017problem, pierson2018fast}, originally developed to model censored outcomes in policing settings, to funnel settings with arbitrary numbers of decision stages. Our generative model, also shown in Figure \ref{fig:model_mechanics}, is as follows.

Consider the patient $i$ progressing through the stages of a hospital stay. At each stage $k$, the clinician estimates some \textit{mortality} risk $p_{i,k} = \text{Pr}(y_i=1)$. We model $p_{i,k}$ as a draw from a \textit{distribution} to reflect the fact that the clinician may have access to unobservables that are informative about mortality risk, and relevant to the clinician's decision, but are not recorded in the data (e.g., whether a patient's speech is slurred). A shape parameter $\delta_k$ modulates the uncertainty in $R$, i.e., $p_{i,k} \sim R(\phi_{i,k}, \delta_k)$, where $\phi_{i,k}$ and $\delta_k$ are the mean and shape parameters of $R$, respectively. 

The mean of this distribution, $\phi_{i,k}$, is a learned function of the covariates available at stage $k$, i.e. $\phi_{i,k} = f_\beta(X_{i,k})$, where $\beta$ denotes the learnable parameters. The parameters $\beta$ may be (partially or fully) shared from stage to stage, or learned separately at each stage. In this paper, we instantiate $f_\beta$ as a logistic model, i.e. $\phi(X) = f_\beta(X) = \text{sigmoid}(\alpha + X^T \beta)$, where $\alpha$ is also learnable.

The clinician then makes a decision on where to send the patient based on the patient's assessed mortality risk $p_{i,k}$. If $p_{i,k}$ exceeds a decision threshold $t_k \in [0,1]$, the clinician admits the patient to the next stage (e.g., the hospital). If it is smaller than this threshold, the patient is discharged. Patients can skip stages if the estimated mortality risk is sufficiently large; for instance, if $p_{i,k}$ exceeds $t_{k+1}$, the subsequent threshold, the clinician may directly admit the patient to a later stage (e.g., ICU), skipping the earlier stage (hospital). Let $D_{i,k}$ be the decision on where to send patient $i$ after stage $k$, as defined in the observed quantities section. Then, we define the decision rule as

\[
D_{i,k} =
\begin{cases}
\text{discharge}, & p_{i,k} < t_k, \\
k+1, & t_k \le p_{i,k} < t_{k+1}, \\
k+2, & t_{k+1} \le p_{i,k} < t_{k+2}, \\
\vdots & \vdots \\
K-1, & t_{K-2} \le p_{i,k} < t_{K-1}, \\
K,   & p_{i,k} \ge t_{K-1}.
\end{cases}
\]

We define the \textit{admit rate} $A_{i,k \rightarrow m}$ for individual $i$ from stage $k$ to some stage $m > k$ (i.e., the rate at which they're sent to that stage) as:
\begin{align}
    A_{i, k \rightarrow m} = Pr(t_{m-1} \leq p_{i,k} < t_m) \label{eq:search_rate}
\end{align}

In our model, the decision $D_{i,k}$ is thus sampled from a categorical distribution:
\begin{align}
    D_{i,k} \sim \text{Cat}(A_{i, k \rightarrow \text{discharge}}, A_{i, k \rightarrow k+1}, \ldots, A_{i, k \rightarrow K}) \label{eq:decision}
\end{align}
If $Y_i$ is eventually observed (as the patient eventually moves to some stage $k > S)$, we additionally model the \textit{mortality rate} as:
\begin{align}
   M_{i,k} = E(Y_i|p_{i,k} < t_k) \label{eq:hit_rate}
\end{align}

\textbf{Implementation details.} We instantiate this model in our MIMIC case study with $K=3$ stages, with $t_1 = t_{hosp}$, or the threshold to hospitalize a patient, and $t_2 = t_{ICU}$, or the threshold to admit a patient to the ICU. The risk distribution $R$ is parameterized using the parameterization from \citet{pierson2018fast}. In particular, we fix the shape parameter $\delta_k$ to be constant across all individuals and let the mean $\phi_{i,k}$ be parameterized with a logistic function. We provide details on its parameters and our inference procedure in Appendix \ref{apd:modeldetails}. We implement our model in Stan and use MCMC to fit all parameters. We maximize the log-likelihood of the observed data, with likelihood contributions from Equation \ref{eq:decision} and \ref{eq:hit_rate}.

\section{Results}
\label{sec:results}
\begin{table}[t]
\centering
\small
\setlength{\tabcolsep}{2pt}
\renewcommand{\arraystretch}{0.95}
\begin{tabular}{@{}p{0.6\columnwidth}cc@{}}
\hline
\textbf{Method} & \textbf{Parameter} & \textbf{MAE} \\
\hline
Logistic Regression (target: ICU)     & $\alpha$      & 2.45 \\
Logistic Regression (target: $y$)     & $\alpha$      & 0.69 \\
Logistic Regression (imputed $y$)     & $\alpha$      & 1.74 \\
\textbf{Funnel Model}                 & $\alpha$      & \textbf{0.16} \\
\hdashline
Logistic Regression (target: ICU)     & $\beta_{1:6}$ & 1.33 \\
Logistic Regression (target: $y$)     & $\beta_{1:6}$ & 0.21 \\
Logistic Regression (imputed $y$)     & $\beta_{1:6}$ & 0.51 \\
\textbf{Funnel Model}                 & $\beta_{1:6}$ & \textbf{0.06} \\
\hline
\end{tabular}
\caption{The funnel model outperforms all logistic regression baselines in estimating the true parameters, achieving the lowest mean absolute error (MAE) across both the intercept $(\alpha)$ and the regression coefficients $(\beta_{1:6})$. Results are shown for three logistic baselines: \emph{Logistic Regression (target: ICU)}, which regresses the decision to admit to the ICU on $X$; \emph{Logistic Regression (target: $y$)}, which regresses $y$ on $X$ for observed $y$; and \emph{Logistic Regression (imputed $y$)}, which regresses $y$ on $X$ after imputing unobserved outcomes.}
\label{tab:aggregated_summary}
\end{table}

\subsection{Synthetic simulations}
\label{sec:synthetic}
We begin by showing that our model can (1) recover the true parameters more accurately, and (2) predict mortality risk better than baselines in synthetic settings where the funnel model is well-specified. To do so, we conduct a synthetic experiment with $K=3$ stages, where certain covariates are revealed only after the first stage, mirroring the setup of MIMIC.

\textbf{Synthetic data generating process.}
We generate data for each patient by sampling six features $X_1, \ldots, X_6 \overset{iid}{\sim} \mathcal{N}(0,1)$ and adding an intercept term. The first three covariates are collected at stage 1 and the final three at stage 2. We sample coefficients $\beta_1, \ldots, \beta_6 \overset{iid}{\sim} \mathcal{N}(0,1)$ and an intercept $\alpha \sim \mathcal{N}(0,1)$ which map $X_{1:6}$ to the mean $\phi$ of the risk distribution $R$. We additionally sample two decision thresholds $t_1, t_2$ and risk distribution shape parameters $\delta_1, \delta_2$.

To generate the observed admissions decisions $D_{i,k}$ and mortality $Y_i$, at each stage $k$, we set $\phi_{i,k} = \text{sigmoid}(\alpha + \beta^T X)$ using the features $X$ available at that stage. We sample $p_{i,k} \sim R(\phi_{i,k}, \delta_k)$, and if this value exceeds $t_k$, we move the patient onto the next stage. We use the final $p_{i,k}$ available for the patient to generate the true $y_i$. That is, we sample the true label $y$ is sampled as $y_i \sim Bern(p_{i,k})$, where $p_{i,k} \sim R(\phi_{i,k}, \delta_k)$ for the last stage $k$ from which the covariates are collected. This $y_i$ is observed if the last stage $k\geq 2$ but unobserved if $k=1$ (i.e., patient is discharged at stage 1). 



\textbf{The funnel model recovers true parameters better than logistic baselines.}
We first assess how other baselines recover the true data-generating coefficients compared to our model. To assess this, we fit three logistic regression baselines, as our data generating process is also a logistic model. 

Our first baseline fits $y$ on all collected features $X$, which we denote as target $y$, except \textit{exclusively} on patients where $y$ is available. Our second baseline fits a different binary variable, whether the patient is admitted to the ICU, on the same $X$. This second baseline uses a mis-specified target (ICU) but is able to use all of the available data. We denote this baseline as target ICU. Our third baseline imputes $y=0$ among the censored population then fits $y$ on $X$ across the entire dataset. We note this baseline, common in prior work on selective labels \citep{chang2024biased}, as imputed $y$. We show that by failing to account for the censoring in the regression, these baselines produce worse model fits.

We track the mean absolute error (MAE) in estimating each of the fitted parameters. We see, as in Table \ref{tab:aggregated_summary}, that the funnel model outperforms logistic regression baselines that do not account for the censoring. Averaged over 1000 simulations, the funnel model improves MAE in estimating the intercept 4-fold and in estimating the regression coefficients 3-fold, compared to the best logistic regression baseline. Appendix \ref{apd:syntheticsims} additionally shows that the funnel model's estimates of the thresholds and shape parameters are well-calibrated.

\begin{table}[t]
\centering
\small
\setlength{\tabcolsep}{3pt}
\renewcommand{\arraystretch}{0.95}
\begin{tabular}{@{}lll@{}}
\hline
Method & AUROC $\uparrow$ & ECE $\downarrow$ \\
\hline
Random Forest                        & 0.625 & 0.347 \\
Logistic Regression (target: ICU)     & 0.743 & 0.155 \\
Logistic Regression (target: $y$)     & 0.722 & 0.302 \\
Logistic Regression (imputed $y$)     & 0.739 & 0.166 \\
\textbf{Funnel Model}                 & \textbf{0.786} & \textbf{0.040} \\
\hline
\end{tabular}
\caption{On synthetic data, the funnel model achieves higher AUROC and lower expected calibration error (ECE) than logistic regression and random forest baselines. We additionally include a non-linear baseline here that does not directly provide feature coefficients.}
\label{tab:auc_ece_synth}
\end{table}

\textbf{The funnel model more accurately predicts true mortality risk.}
Next we assess the funnel model's ability to predict true mortality risk on the censored population. In particular, we show that the funnel model is able to better predict the censored $y$ compared to other baselines which do not explicitly account for the censoring.

To simulate this setting, we generate $y$ for all patients and then manually censor it for patients who do not make it to the last stage (ICU). We then assess our model (and baseline models') accuracy in predicting risk on the censored population.

In this setting, our performance metric is predictive ability (i.e., AUROC, ECE), so we include a non-linear baseline, random forest, which does not provide interpretable regression coefficients but can still make predictions on the censored population. We find in Table \ref{tab:auc_ece_synth} that the funnel model is able to generalize better to the censored population, both in predictive power (AUROC) and in calibration (ECE).

\subsection{MIMIC Data and Model Setup}\label{sec:mimic_data}
For our real-world case study, we model mortality risk as patients flow from the emergency department (ED) to the hospital and ICU. We use the MIMIC-IV dataset for this case study. The MIMIC-IV dataset consists of the Beth-Israel Deaconness Medical Center's (BIDMC) de-identified electronic health records from January 2008 to December 2019. The BIDMC is composed of multiple hospital wards, including multiple ED and ICU wards. Patients can be discharged from the BIDMC from the ED, hospital, or ICU. Death at discharge, encoded as a binary variable, is unobserved for ED patients but is observed for hospital and ICU discharged. We include two threshold parameters in our model: (1) $t_{hosp}$, the mortality risk threshold to hospitalize a patient, and (2) $t_{ICU}$, the mortality risk threshold to admit a patient to the ICU. We constrain $t_{hosp} < t_{ICU}$ to reflect that the ICU is typically reserved for more acute cases.

We include 25 covariates, based on features collected at triage. We $z$-score the numeric features and include a quadratic term, the square of the $z$-score, to account for non-linearities in the relationship between triage features and mortality risk (e.g., very high and very low high heart rates may both indicate high risk). We additionally one-hot encode the chief complaint (e.g. dizziness or shortness of breath). Once the patient is hospitalized, we collect their age, which is included as an additional covariate from the hospital stage onwards. Details on the MIMIC dataset are provided in Appendix \ref{apd:mimicdetails}.

\subsection{Case study: Gender disparities in MIMIC}\label{sec:gender}


\begin{table}[t]
\centering
\small
\setlength{\tabcolsep}{3pt}
\renewcommand{\arraystretch}{0.95}
\begin{tabular}{@{}p{0.46\columnwidth}ccc@{}}
\hline
\textbf{Method} & \textbf{Gender} & \textbf{AUROC} $\uparrow$ & \textbf{ECE} $\downarrow$ \\
\hline
Random Forest                                      & F & 0.535 & 0.285 \\
\shortstack[l]{Logistic Regression\\(target: ICU)} & F & 0.674 & 0.276 \\
\shortstack[l]{Logistic Regression\\(target: $y$)} & F & 0.658 & 0.145 \\
\shortstack[l]{Logistic Regression\\(imputed $y$)} & F & 0.633 & 0.472 \\
\textbf{Funnel Model}                              & F & \textbf{0.677} & \textbf{0.011} \\
\hdashline
Random Forest                                      & M & 0.527 & 0.256 \\
\shortstack[l]{Logistic Regression\\(target: ICU)} & M & 0.669 & 0.331 \\
\shortstack[l]{Logistic Regression\\(target: $y$)} & M & 0.643 & 0.108 \\
\shortstack[l]{Logistic Regression\\(imputed $y$)} & M & 0.600 & 0.461 \\
\textbf{Funnel Model}                              & M & \textbf{0.678} & \textbf{0.014} \\
\hline
\end{tabular}
\caption{Averaged AUROC and ECE on MIMIC data, comparing the funnel model against logistic regression and random forest baselines. Metrics are averaged across ICU admission and mortality prediction tasks. Mortality is censored for ED patients, so mortality prediction is evaluated on hospitalized patients only. The funnel model achieves superior predictive performance and calibration across both genders.}
\label{tab:auc_ece_by_gender_method}
\end{table}

\begin{figure}[h] 
\centering
\setlength{\belowcaptionskip}{-2em} 
\includegraphics[width=0.5\textwidth]{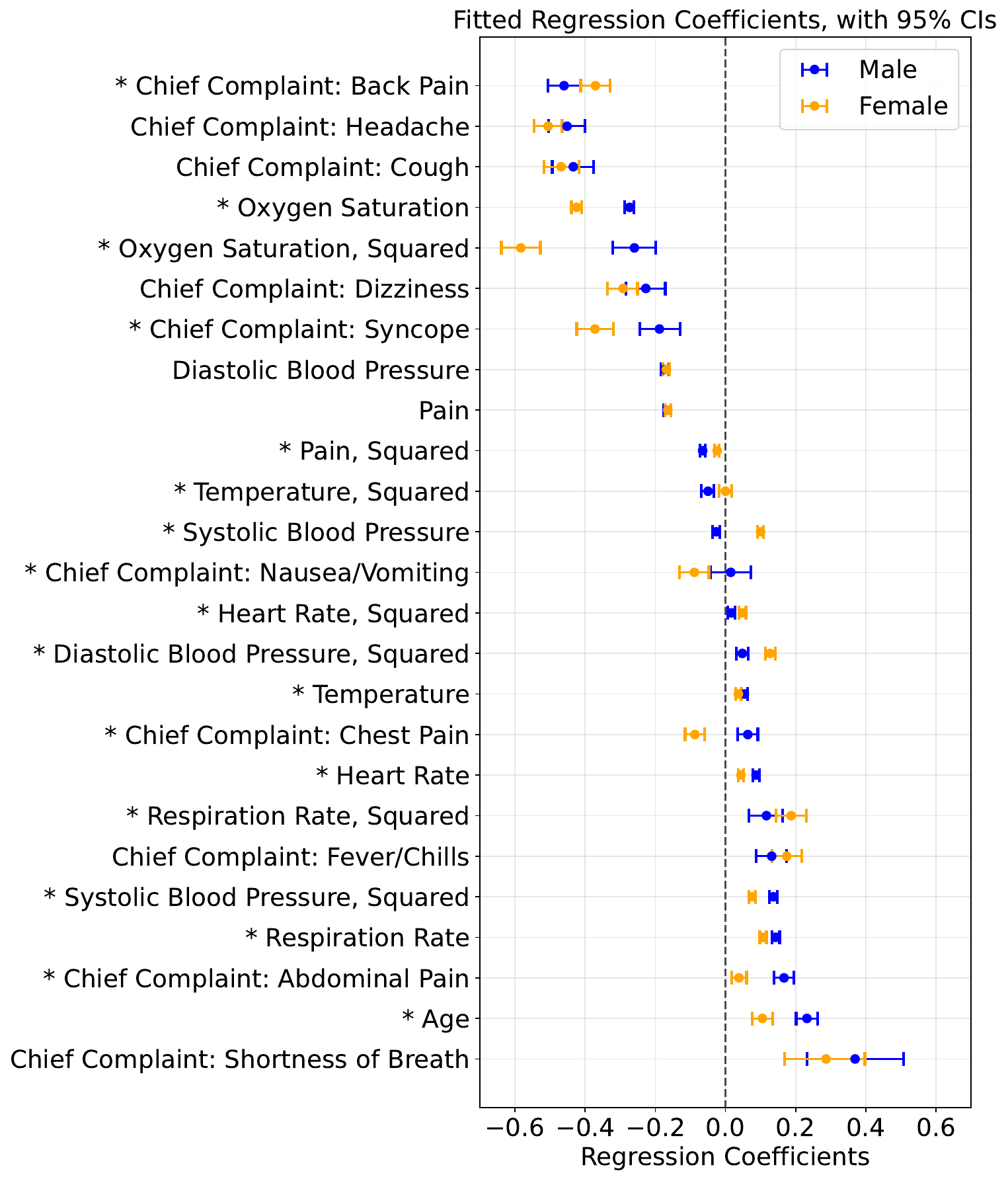}
\caption{Fitted coefficients in MIMIC for male (blue) and female (yellow) patients. Asterisks indicate statistically significant differences by gender in the fitted coefficients. There are notable differences in the relative importance of age and O2 saturation in predicting mortality risk across gender.}
\label{fig:gender_results}
\end{figure}

We apply our model to analyze how emergency department decisions in MIMIC differ across male and female patients.\footnote{We follow language from MIMIC in describing the gender column name and values. Specifically, our analysis uses the `gender' variable from MIMIC-IV-ED's table, which is described in MIMIC as ``The patient's administrative gender as documented in the hospital system"; within that column, 229,898 patients are charted as ``F" and 195,189 patients are charted as ``M".} To do so, we fit our funnel model to male and female patients in MIMIC separately, using the set of covariates and pathways described in Section \ref{sec:mimic_data}. In total, our dataset contains 229,898 and 195,189 unique visits from female and male patients, respectively.

\textbf{The funnel model outperforms baselines in real-world MIMIC data.}
We begin by assessing whether our model can outperform the baselines we introduced in Section \ref{sec:synthetic} in our real-world case study. Mortality records (i.e., $y$) are censored in MIMIC for patients discharged from the ED. We instead define two prediction tasks from the data we have: (1) predicting mortality among patients whose mortality is observed, and (2) predicting ICU admission. Our trained funnel model directly provides estimates of the admit rates and mortality rates for every patient. Given this is a multi-task prediction problem, we report metrics averaged across both tasks. 

To compare these predictions to baselines, we compare to logistic regression and random forest baselines and assess their predictive performance (average AUROC across the two tasks) and calibration. Table \ref{tab:auc_ece_by_gender_method} compares the funnel model to other baselines, all of which have worse AUROC than the funnel model. The random forest baseline performs well in-distribution (i.e., on the training target, $y$ in ICU), but fails to generalize to the out-of-distribution (censored $y$ and ICU admission) tasks, so the averaged AUROC is close to random. There are additionally substantial improvements in calibration, with the funnel model always achieving an ECE less than 0.02, while the next-best baseline has a calibration error of 0.10. There are cases where the performance of one of these models is better at predicting one particular outcome. For instance, models trained to \textit{only} predict $y$ tend to have better AUROC in predicting $y$, at the cost of having lower AUROC in predicting ICU admissions. By contrast, the funnel model is able to predict both well, as shown in the appendix.
In addition, we demonstrate that our model's posterior estimates of the true admit and mortality rates are accurate, which we provide in Appendix \ref{apd:otherresults}.

\textbf{Gender differences in admissions thresholds.} We find that the estimated mortality risk thresholds $t_{hosp}$ and $t_{ICU}$ for admitting patients to the hospital and ICU, respectively, is higher for female patients than male patients, indicating that female patients must be at greater risk of dying to be admitted to the hospital or ICU than male patients. Our model estimates that the mortality risk threshold to admit male patients to the hospital is $t_{hosp, M} = 0.0032$ (95\% CI: $0.0028- 0.0036$), 25\% lower (in relative terms) than that for female patients, $t_{hosp, F} = 0.0043$ (95\% CI: $0.0039- 0.0047$). Similarly, the estimated threshold to admit male patients to the ICU, $t_{ICU, M} = 0.045$ (95\% CI: $0.043- 0.047$), is 10\% lower than the threshold to admit female patients $t_{ICU, F} = 0.051$ (95\% CI: $0.048- 0.053$). Both differences are statistically significant: the 95\% confidence interval on $t_{hosp, F} - t_{hosp, M}$ is $(0.00055, 0.0017)$ and the 95\% confidence interval on $t_{icu, F} - t_{icu, M}$ is $(0.002, 0.008)$. Female patients are admitted to the hospital and ICU at lower rates (46\% and 5.7\%, respectively) than male patients (52\% and 8.0\%, respectively) but die at comparable or higher rates (13.1\% ICU mortality for women, 12.4\% ICU mortality for men; 0.5\% hospital mortality for women, 0.6\% hospital mortality for men). This indicates that our model's findings are justified within the data.

\textbf{Gender differences in triage acuity assessment.}
We next examine whether clinician assessed risk at triage differs across gender. We do this by examining how our model's estimated mortality risk compares to the triage acuity scores. In MIMIC, acuity is scored from 1 (most severe) to 5 (least severe), so in general we should expect patients with higher estimated mortality risk to have lower acuity scores.

To evaluate gender differences in triage acuity, we run a risk-adjusted regression to see if gender provides any additional signal in predicting acuity. In particular, we fit the regression 

$$\text{acuity}_i = \beta_M \cdot \text{male}_i + \beta_F \cdot \text{female}_i + \beta_{\text{risk}} \cdot \text{risk}_i + \epsilon_i$$

where $\text{risk}_i$ is the model-estimated risk for patient $i$, $\text{male}_i$ is a binary indicator whether the patient is male, $\text{female}_i$ is a binary indicator whether the patient is female, and $\epsilon_i \sim \mathcal{N}(0,1)$ is a noise term. We find that $\beta_F$ is larger (2.967, CI: [2.962, 2.971]) than $\beta_M$ (2.940, CI: [2.935, 2.945]), indicating that women are assessed to have less acute cases at the same estimated mortality risk, although the magnitude of this difference is small.

\textbf{Discussion of gender disparities in thresholds and acuity.}
Although these results indicate gender disparities, it is important to note our model does not explain \textit{why} those disparities might exist. Moreover, these differences, while statistically significant, are small in magnitude. In particular, our results should not necessarily be taken to imply that clinicians have gender-based biases, though that is one possibility. For example, one non-bias explanation would be that thresholds are time-varying (e.g., higher thresholds to admit to ICU when ICU is near capacity), and female patients are more likely to come to the ED at capacity-constrained times (data which is unavailable in MIMIC). Other possible confounders include insurance (e.g., women are uninsured at higher rates) or comorbidities (e.g., women have higher rates of comorbitities unaccounted by our model).

Beyond gender, our approach can be used to analyze differences in emergency care across other features, including insurance status, marital status, and race. We provide results from applying the same model to different race groups in MIMIC in the Appendix.

\textbf{Gender differences in regression coefficients.}
Finally, we investigate how risk differs across gender by analyzing the fitted coefficients in our regression models. Figure \ref{fig:gender_results} shows how the estimated regression coefficients differ across male and female patients. In general, the most important predictors of mortality risk include chief complaints including shortness of breath, back pain, and headache; age; and oxygen saturation, consistent with prior work \citep{fernandes2020predicting, stretch2021criteria}. For instance, our model estimates that the mortality odds increase by a factor of 2.2 if a (male) patient's chief complaint is shortness of breath, as opposed to back pain.

Although the coefficients generally agree across gender, there are some notable differences. For instance, age increases mortality risk among male patients (coefficient: 0.232) more significantly than female patients (coefficient: 0.105).  Conversely, oxygen saturation is far more important in predicting female patients' mortality risk (coefficient: -0.424) than male patients' (coefficient: -0.273), consistent with prior work on gender-based differences in hypoxia \citep{jun2021analysis}. A few variables, such as diastolic blood pressure, have negative coefficients for the linear term, indicating that smaller values are generally worse, but also have positive coefficients for the quadratic term, indicating that large deviations from the mean (e.g. very high blood pressure) also increase mortality risk.

\section{Discussion}\label{sec:discussion}
Our work demonstrates the importance of modeling funnel decision structures, where labels are censored as a result of a sequence of human decisions. By modeling both the label and the censoring decisions jointly, we are able to more accurately capture the true parameters and predict mortality and admissions decisions than models which fail to account for the censoring. Applying our work to MIMIC, we find evidence of gender-based disparities in emergency care, with women facing slightly higher thresholds for ICU admission than men. Our model does not explain the sources of this disparity, and we leave it as future work to more carefully investigate the sources. Other future work includes using time-varying thresholds (e.g., admission decisions change when ICU is at capacity), examining other disparities (e.g., by insurance status), and modeling other funnel decisions, potentially beyond healthcare settings too.

\section{Acknowledgements}
We thank Sidhika Balachandar, Divya Shanmugam,  Raj Movva, Drew Prinster, and Gabriel Agostini for helpful comments and feedback on the work. This work was supported by a Hertz Fellowship, NSF Graduate Research Fellowship under grant number 2141064, NIH R35 GM141861-05, a Google Research Scholar award, an AI2050 Early Career Fellowship, NSF CAREER \#2142419, a CIFAR Azrieli Global scholarship, a gift to the LinkedIn-Cornell Bowers CIS Strategic Partnership, the Survival and Flourishing Fund, Open Philanthropy, and the Zhang Family Endowed professorship.

\bibliography{jmlr-sample}

\appendix

\section{Model Details}\label{apd:modeldetails}

\subsection{Risk distribution parameterization}\label{apd:risk}
Our goal is to write a distribution $R$, defined on the unit interval $[0,1]$, such that if $p \sim R$, it is easy (and fast) to compute the admit and mortality rates, as defined in Equations \ref{eq:search_rate} and \ref{eq:hit_rate}, respectively. A natural choice for $R$ might be a beta distribution. However, it is expensive to compute $Pr(p > t)$ and $E_{p \sim R}[p|p>t]$ and associated derivatives if $R$ is a beta distribution \citep{boik1999derivatives}. Discriminant distributions, as introduced in \citep{pierson2018fast}, offer a scalable workaround.

Let $Y \sim \text{Bern}(p)$. We assume the distribution of some signal $X$ (e.g., a learned function of the features) is distributed $X | Y=0 \sim \mathcal{N}(0, \sigma)$ and $X | Y=1 \sim \mathcal{N}(\mu, \sigma)$. Define $g(x) = Pr(Y=1|X=x)$. Prior work \citep{pierson2018fast} shows that $g$ is a monotonic function of $x$.

If we define $\delta = \frac{\mu}{\sigma}$, then it is possible to parameterize $R$ in terms of $\phi$, the mean parameter, and $\delta$, a shape parameter, alone. Moreover:

\begin{align}
    g(x) = \frac{1}{1 + \frac{1 - \phi}{\phi} \exp(-\delta x + \delta^2/2)}
\end{align}

This yields exact analytic expressions for the admit and mortality rates. In particular:

\begin{align*}
    Pr(p > t) &= (1 - \phi) (1 - \Phi(g^{-1}(t); 0,1)) \\
    &+ \phi (1 - \Phi(g^{-1}(t); \delta,1)) \\
    E[Y|p > t] &= \frac{\phi (1 - \Phi(g^{-1}(t); \delta,1))}{Pr(p>t)}
\end{align*}

where $\Phi$ is the normal CDF. 

We parameterize $\phi$ as a learned (logistic) function of covariates. In particular, we set $\phi_{i,k} = f_\beta(X_{i,k})$, for the $i$th individual at stage $k$, and $f_\beta(X) = \text{sigmoid}(\alpha + X^T \beta)$, where the intercept $\alpha$ is also a learnable parameter. Although we parameterize this with a logistic function, other parameterizations are also possible. For instance, one could set $f_\beta$ to a small neural network.

\subsection{List of MIMIC pathways and corresponding likelihood contributions}

There are four possible pathways a patient may take in this MIMIC setup: (1) discharged from ED, (2) admitted from ED to hospital to ICU,
(3) admitted from ED to hospital, then discharged,  (4) admitted from ED to ICU.

We write out the likelihood contributions for each of these four.

\textbf{ED discharge} If the patient is discharged from the ED, their only likelihood contribution is via the admit rate:
$$A_{i, ED \rightarrow discharge} = Pr(p_{i, ED} < t_{hosp})$$

\textbf{ED, hosp, ICU}
The admit rate from the ED to the hospital will be:
$$A_{i, ED \rightarrow hosp} = Pr(t_{hosp} \leq p_{i, ED} < t_{ICU})$$

This patient's risk is then updated based on the new covariates collected in the hospital to $p_{i, hosp}$. Afterwards the clinician decides whether to admit this patient to the ICU, and the corresponding admit rate is:
$$A_{i, hosp \rightarrow ICU} = Pr(t_{ICU} \leq p_{i, hosp})$$

For patients who reach the ICU, we compute the mortality rate as:
$$E(Y_i|p_{i, hosp} > t_{ICU})$$

\textbf{ED, hosp, discharge}
The admit rate from the ED to the hospital is computed the same as above. We update the patient's risk in the hospital to $p_{i, hosp}$ based on the covariates collected in the hospital. The admit rate for the discharge is:

$$A_{i, hosp \rightarrow discharge} = Pr(p_{i, hosp} < t_{ICU})$$

The corresponding (and distribution-matched) mortality rate is:
$$M_{i, hosp \rightarrow discharge} = E(Y_i|p_{i, hosp} < t_{ICU})$$

Note that this mortality rate matches the most recent admit rate. The most recent admit rate is the rate of discharge from the hospital, and this mortality rate matches that admit rate.

\textbf{ED, ICU}
The admit rate for ED to ICU is:
$$A_{i, ED \rightarrow ICU} = Pr(p_{i, ED} \geq t_{ICU})$$

We then calculate the mortality rate for these patients as:
$$M_{i, ED \rightarrow ICU} = Pr(Y_i|p_{i, ED} \geq t_{ICU})$$

Putting together these with \ref{apd:risk}, we compute the admit and mortality rates in MIMIC as (shortening $S_{i, ED \rightarrow hosp}$ to $S_{i, e \rightarrow h}$):

\begin{align*}
    A_{i, e \rightarrow h} &= Pr(t_{hosp} \leq p_{i,ED} < t_{ICU}) \\
    &= \phi_{i,ED}\left[\Phi\left(g^{-1}\left(t_{ICU}\right) - g^{-1}(t_{hosp}); \delta_{hosp}, 1\right) \right] \\
    &+ (1-\phi_{i,ED})\left[\Phi\left(g^{-1}\left(t_{ICU}\right) - g^{-1}(t_{hosp}); 0, 1\right) \right]
\end{align*}

The mortality rate is only computed after failing (or passing) a particular stage $k$ ($k=$ICU or hosp), and thus can be calculated as:

$$H_{i,k} = \frac{\phi_{i,k} \Phi(g^{-1}(t_k); \delta_k, 1)}{Pr(p_{i,k} < t_k)}$$

Each of these terms $S_{i, .}, H_{i, .}$ contributes to the model's likelihood during fitting. For each patient, we observe a flow through the hospital and, for a subset of patients, the true label $y$, and learn the parameters $\beta$ that maximize the likelihood of the observed pathways and $y$.

\subsection{Model fitting details}
We set priors on all parameters. For the intercept, we set $\alpha \sim \mathcal{N}(0,1)$, although we use $\alpha \sim \mathcal{N}(-5,1)$ in our MIMIC experiments to reflect the fact that the baseline mortality rate is very small (1\%). All the other regression coefficients have priors $\beta_i \sim \mathcal{N}(0,1)$.

Our threshold priors are $t_{hosp} \sim \text{Half-Normal}(0, 0.5)$, $t_{ICU} \sim t_{hosp} + \text{Half-Normal}(0, 0.5)$. We constrain the smaller threshold to be in the interval $[0, 0.5]$ and the larger threshold to be in the interval $[0,1]$. The shape parameter priors are also $\delta_1 \sim \text{Half-Normal}(0, 0.5)$ and $\delta_2 \sim \text{Half-Normal}(0, 0.5)$. 

We implemented our model in Stan and interfaced with it in cmdstanpy, which enabled GPU support. We ran all experiments shown in the paper on a single NVIDIA A6000 48GB GPU. We fit our model using MCMC using 500 warmup steps and 500 sampling steps. To ensure convergence, we verified that $\hat{R} \leq 1.05$ for all parameters on our real-data experiments.

\section{Synthetic Simulations}\label{apd:syntheticsims}
\begin{figure}[h]
\begin{center}
\includegraphics[width=0.5\textwidth]{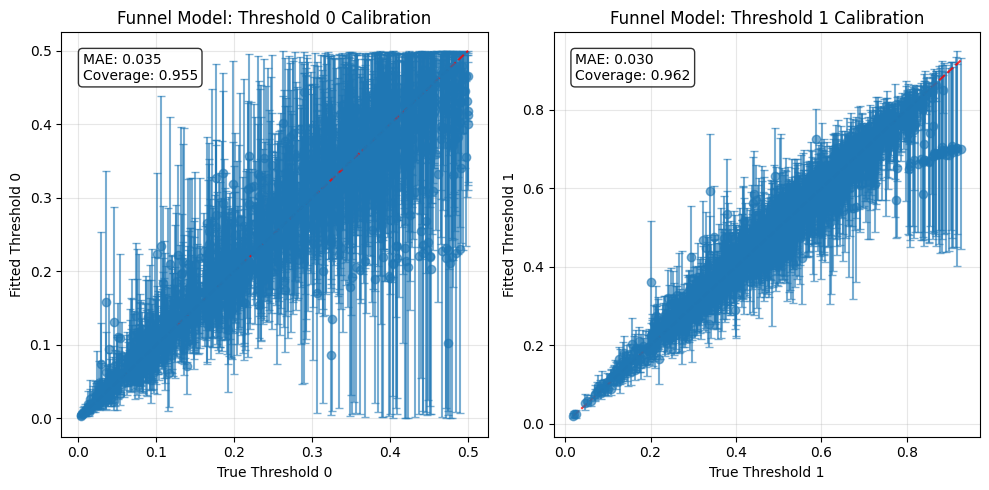}
\caption{Calibration plots of threshold parameters, funnel model.}
  \label{fig:threshold_calibration}
\end{center}
\end{figure}

\begin{figure*}
\begin{center}
\includegraphics[width=\textwidth]{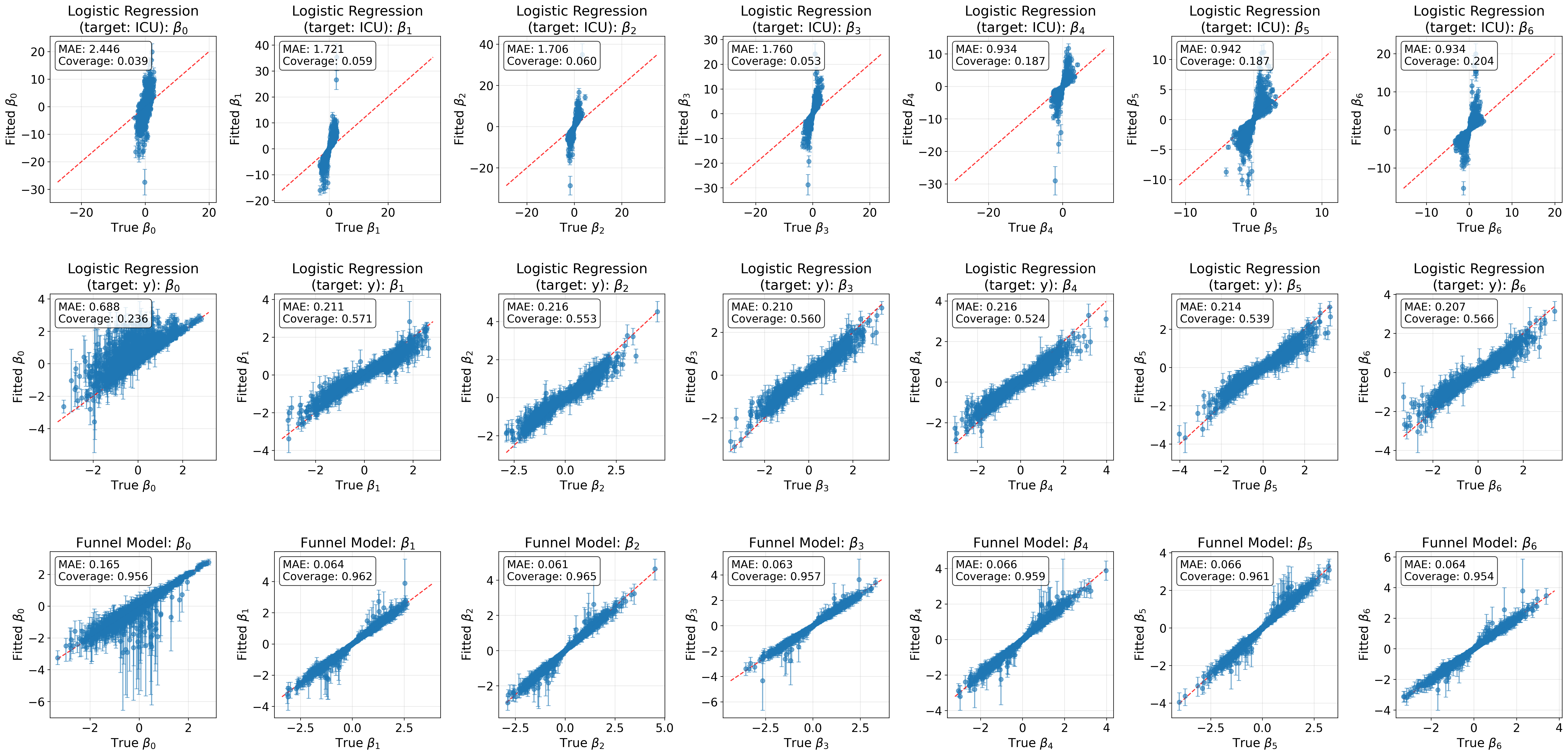}
\caption{Calibration plots of logistic regression baselines and funnel model. Coverage is the empirical coverage of the 95\% confidence intervals. }
  \label{fig:calibration_plots}
\end{center}
\end{figure*}

\subsection{Synthetic data details}

As noted in \ref{sec:synthetic}, we sample $X_1, \ldots, X_6 \overset{iid}{\sim} \mathcal{N}(0,1)$ and $\alpha, \ldots, \beta_6 \overset{iid}{\sim} \mathcal{N}(0,1)$. At each stage $k$ ($k=1$ or 2), we set $\phi_{i,k} = \text{sigmoid}(\alpha + \beta_{1:6}^T X_{1:6})$. We sample $p_{i,k} \sim R(\phi_{i,k}, \delta_k)$, and if this value exceeds $t_k$, we move the patient onto the next stage.

We sample $t_1 \sim \text{Unif}(0, 0.5)$ and $t_2 \sim t_1 + \text{Unif}(0, 0.5)$ and $\delta_1 \sim \text{Half-Normal}(0,0.5)$, $\delta_2 \sim \text{Half-Normal}(0,0.5)$.

We parameterize the risk distribution mean with a logistic model. $X_1, X_2, X_3$ are revealed at stage 1 (ED), so $\phi_{i,1} = \text{sigmoid} (\alpha + \beta_{1:3}^TX_{1:3})$. We then sample $p_{i,1} \sim R(\phi_{i,1}, \delta_1)$. If $p_{i,1} > t_1$, then patient $i$ passes stage 1 (i.e., admitted to the hospital) and three more features $X_4, X_5, X_6$ are revealed. We then repeat: $\phi_{i,2} = \text{sigmoid} (\alpha + \beta_{1:6}^T X_{1:6})$. If the sampled $p_{i,2} > t_2$, the patient makes it to the last stage (i.e., admitted to ICU), and the true label $y$ is sampled as $y \sim Bern(p_{i,2})$, where $p_{i,2} \sim R(\phi_{i,2}, \delta_2)$.

For each simulation, we ensure that the distribution of stages and outcomes is reasonable (e.g., avoiding pathological cases such as all patients end up in ICU). In particular, we only keep simulations where at least ten patients are observed in each pathway.

\subsection{Baseline details}
We compare against three sets of baselines, explained in \ref{sec:synthetic}. For the logistic regression baselines, we train on model $y \sim X$ only on the population where $y$ and $X$ is observed. We additionally include another baseline $ICU \sim X$, which enables us to use all the samples, at the cost of mis-specifying the target ($ICU$ instead of $y$). Lastly, we train a model $y \sim X$ where we impute $y=0$ for the missing population. Figure \ref{fig:calibration_plots} shows that calibration of the first two baselines' fitted coefficients across several simulations. We additionally plot our model's fitted coefficients in this plot.

\newpage

\newpage
\section{Additional results}\label{apd:otherresults}

\subsection{Posterior predictive checks}
\begin{figure}[h] 
\centering
\setlength{\belowcaptionskip}{-2em} 
\includegraphics[width=0.45\textwidth]{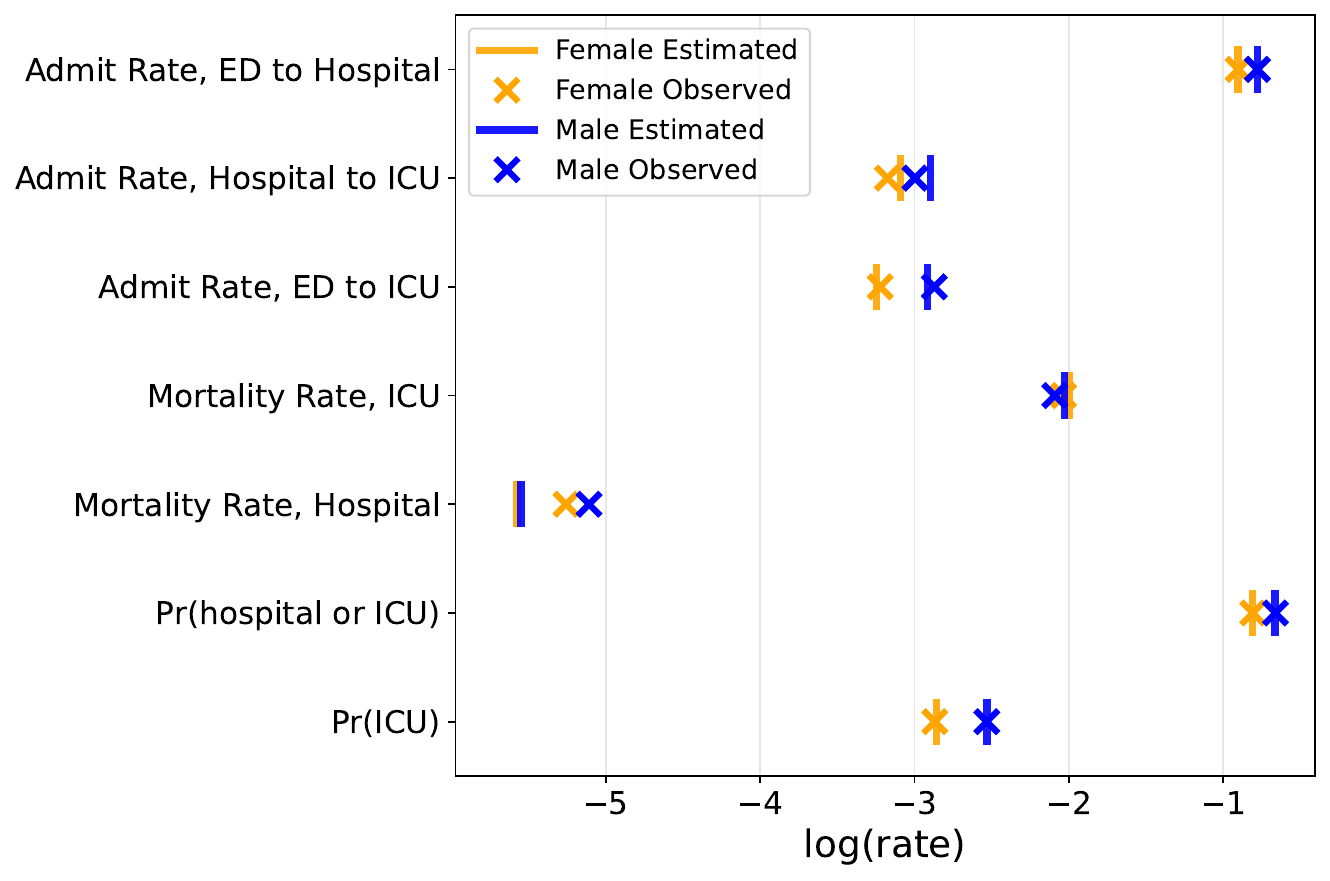}
\caption{ Our model is able to recover the true admit and mortality rates in MIMIC data.}
\label{fig:admitrates}
\end{figure}

 Using posterior predictive checks (i.e., inferred quantities from posterior), we additionally validate that our model is able to recover the admit and mortality rates in MIMIC, which we show in Figure \ref{fig:admitrates}.

 \subsection{Racial subgroup results}
 We additionally apply our model separately to each coarse race subgroup in MIMIC \citep{movva2023coarse}. We omit ``Native American" as a racial category due to small sample size and ``Other" due to lack of specificity.

\begin{figure}[h] 
\centering
\setlength{\belowcaptionskip}{-2em} 
\includegraphics[width=0.5\textwidth]{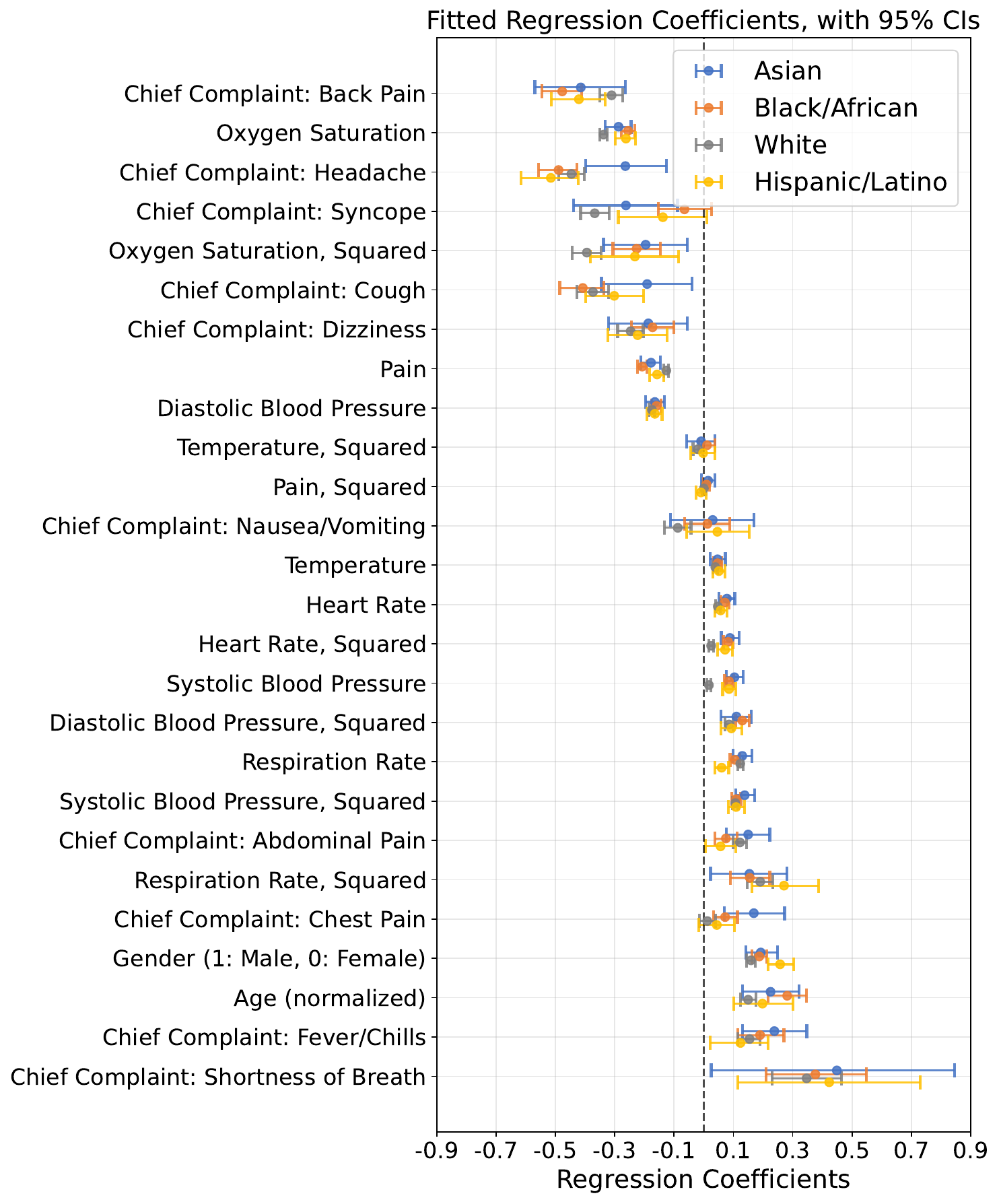}
\caption{The fitted coefficients in MIMIC for each race.}
\label{fig:race}
\end{figure}

 Table \ref{tab:race_thresholds} shows the fitted thresholds by race. In general, we observe that Asian patients have the highest mortality risk thresholds for both hospital and ICU admission. This aligns with observations that their mortality rates in the ICU and hospital are higher than any other group, despite having comparable admit rates. Figure \ref{fig:race} further examines race-specific coefficients in MIMIC.


\begin{table*}[t]
\centering
\small
\setlength{\tabcolsep}{5pt}
\renewcommand{\arraystretch}{0.95}
\begin{tabular}{@{}lcccccc@{}}
\hline
\textbf{Race} & 
\shortstack{Hospitalization\\Threshold} & 
\shortstack{ICU\\Threshold} & 
\shortstack{Hospital\\Mortality} & 
\shortstack{ICU\\Mortality Rate} & 
\shortstack{Hospital\\Admit Rate} & 
\shortstack{ICU\\Admit Rate} \\
\hline
Asian           & 0.008 [0.006, 0.010] & 0.064 [0.055, 0.073] & 0.007 & 0.146 & 0.394 & 0.054 \\
Black/African   & 0.003 [0.003, 0.004] & 0.039 [0.036, 0.042] & 0.004 & 0.097 & 0.395 & 0.043 \\
White           & 0.003 [0.003, 0.004] & 0.047 [0.045, 0.049] & 0.006 & 0.121 & 0.536 & 0.074 \\
Hispanic/Latino & 0.004 [0.003, 0.005] & 0.037 [0.031, 0.043] & 0.003 & 0.084 & 0.363 & 0.037 \\
\hline
\end{tabular}
\caption{Race-specific decision thresholds and outcome rates. Hospitalization and ICU thresholds are estimates from our model (95\% CIs in brackets). The mortality and admission rates are both empirically calculated on the MIMIC data.}
\label{tab:race_thresholds}
\end{table*}

\newpage

\section{MIMIC details}\label{apd:mimicdetails}

\subsection{Data details}
The MIMIC database contains several hundreds of thousands of patients. Each patient is given a unique subject ID. Within each stay at the ED, the patient is given a unique stay ID. Patients are triaged upon arrival at the ED, providing nine common covariates: temperature (\textdegree F), heart rate (bpm), respiratory rate (bpm), oxygen saturation (\%), systolic blood pressure (mmHg), diastolic blood pressure (mmHg), chief complaint, pain, and acuity, which is measured according to the Emergency Severity Index (ESI) Five Level triage system and assigned by a registered nurse. We omit acuity as a predictor in our model, as we use it to correlate our model's estimated mortality risk to the triage acuity.

Missing values are replaced by the average of the non-missing values. Pain, which is right-skewed, is self-reported by the patient on a scale of $0-10$. Values deviating outside the range $0-10$ were replaced with the average. In general, for values outside the range of plausible values, such as body temperatures above 200F and oxygen saturation above 100\, we impute the mean. We $z$-scored all numeric variables to better compare effect sizes. We additionally included a quadratic term, $z^2$ for these variables, as the mortality risk was often non-monotone in the observed value. For instance, very low and very high heart rates are both concerning. We additionally one-hot encoded chief complaint into one of ten possible complaints. Age is collected upon admission to the hospital (or ICU), along with language and marital status. We include age as a predictor in our experiments but omit language and marital status.

We define ICU wards as any hospital ward identified as some sort of ``ICU" or ``Intensive Care Unit."  We consider the universe of all patients who were admitted to the BIDMC through the ED. This leaves 425,087 patients. Of these patients, 182,803 are admitted to the (non-ICU wards of the) hospital, and 20,213 are admitted directly to the ICU. An additional 8,368 patients are admitted from the hospital to the ICU.

\newpage

\begin{table}[t]
\centering
\small
\begin{tabular}{l l l p{0.42\linewidth}}
\hline
\textbf{Feature} & \textbf{Type} & \textbf{First available at} & \textbf{Description} \\
\hline
intercept & Intercept (constant) & ED & Constant 1. \\
temperature & numeric, $Z$-scored & ED & Triage body temperature ($^\circ$F). \\
heartrate & numeric, $Z$-scored & ED & Heart rate (bpm). \\
resprate & numeric, $Z$-scored & ED & Respiratory rate (breaths/min). \\
o2sat & numeric, $Z$-scored & ED & Pulse oximetry SpO$_2$ (\%). \\
sbp & numeric, $Z$-scored & ED & Systolic blood pressure (mmHg). \\
dbp & numeric, $Z$-scored & ED & Diastolic blood pressure (mmHg). \\
pain & numeric, $Z$-scored & ED & Pain score (0--10). \\
temperature\_sq & numeric, $Z$-scored then squared & ED & Square of $Z$-scored temperature. \\
heartrate\_sq & numeric, $Z$-scored then squared & ED & Square of $Z$-scored heart rate. \\
resprate\_sq & numeric, $Z$-scored then squared & ED & Square of $Z$-scored respiratory rate. \\
o2sat\_sq & numeric, $Z$-scored then squared & ED & Square of $Z$-scored SpO$_2$. \\
sbp\_sq & numeric, $Z$-scored then squared & ED & Square of $Z$-scored systolic BP. \\
dbp\_sq & numeric, $Z$-scored then squared & ED & Square of $Z$-scored diastolic BP. \\
pain\_sq & numeric, $Z$-scored then squared & ED & Square of $Z$-scored pain score. \\
chiefcom\_chest\_pain & one-hot & ED & Chief complaint: chest pain. \\
chiefcom\_abdominal\_pain & one-hot & ED & Chief complaint: abdominal pain. \\
chiefcom\_headache & one-hot & ED & Chief complaint: headache. \\
chiefcom\_shortness\_of\_breath & one-hot & ED & Chief complaint: shortness of breath. \\
chiefcom\_back\_pain & one-hot & ED & Chief complaint: back pain. \\
chiefcom\_cough & one-hot & ED & Chief complaint: cough. \\
chiefcom\_nausea\_vomiting & one-hot & ED & Chief complaint: nausea/vomiting. \\
chiefcom\_fever\_chills & one-hot & ED & Chief complaint: fever/chills. \\
chiefcom\_syncope & one-hot & ED & Chief complaint: syncope. \\
chiefcom\_dizziness & one-hot & ED & Chief complaint: dizziness. \\
age & numeric, $Z$-scored & Hospital/ICU & Age (years). \\
\hline
\end{tabular}
\caption{Feature dictionary for MIMIC predictors.}
\label{tab:mimic_features}
\end{table}

\end{document}